\documentclass[conference]{IEEEtran}
\IEEEoverridecommandlockouts
\usepackage{cite}
\usepackage{amsmath,amssymb,amsfonts}
\usepackage{graphicx}
\usepackage{textcomp}
\usepackage{xcolor}

\begin{document}

\title{Anatomical Landmark-Guided Deep Reinforcement Learning for Autonomous Gastric Navigation}

\DeclareRobustCommand*{\IEEEauthorrefmark}[1]{\raisebox{0pt}[0pt][0pt]{\textsuperscript{\footnotesize #1}}}

\author{
    \IEEEauthorblockN{
        Haoxuan Wu\IEEEauthorrefmark,
        Sishen Yuan\IEEEauthorrefmark,
        Haitao Gao\IEEEauthorrefmark,
        Zhen Li\IEEEauthorrefmark,
        Xiuli Zuo\IEEEauthorrefmark,
        Hongliang Ren\IEEEauthorrefmark{*}
    }
    \IEEEauthorblockA{Department of Electronic Engineering, The Chinese University of Hong Kong, Hong Kong SAR, China}
    \thanks{*Corresponding authors: hren@cuhk.edu.hk }
}

\maketitle

\begin{abstract}
Wireless capsule endoscopy (WCE) enables painless visualization of the gastrointestinal tract, but its diagnostic potential is limited by incomplete mucosal coverage and poor transferability of existing navigation methods across patient anatomies. We propose a transferable, anatomical landmark-guided deep reinforcement learning (AL-DRL) framework for autonomous gastric navigation. Leveraging a lightweight edge-contour-depth fusion module, our policy operates on stable, low-dimensional landmark coordinates rather than high-dimensional video streams, effectively bridging the sim-to-real gap. In simulations across eight patient-derived models, the method achieves over 97\% coverage within 50 seconds, significantly outperforming vanilla PPO, SAC, and DQN agents. A two-stage sim-to-real pipeline with an adaptive dynamic programming controller actively mitigates physical disturbances. Ex-vivo experiments demonstrate a mean coverage of 87\% and a 53\% reduction in procedure time compared with expert manual control.
\end{abstract}


\section{Introduction}

Wireless capsule endoscopy (WCE) has revolutionized gastrointestinal diagnostics by enabling minimally invasive visualization of the entire digestive tract~\cite{cao2024}. Maximizing mucosal surface coverage is a central requirement for next-generation WCE systems, yet achieving complete gastric coverage remains challenging due to the absence of fully autonomous navigation~\cite{moglia2009}.

Deep reinforcement learning (DRL) has shown promise for autonomous capsule navigation~\cite{zhang2022robio}, yet existing ``end-to-end'' visual DRL approaches suffer from overfitting to training-environment morphology, high computational demands, and vulnerability to the visual domain gap across patients.

\begin{figure}[!t]
    \centering
    \includegraphics[width=\columnwidth]{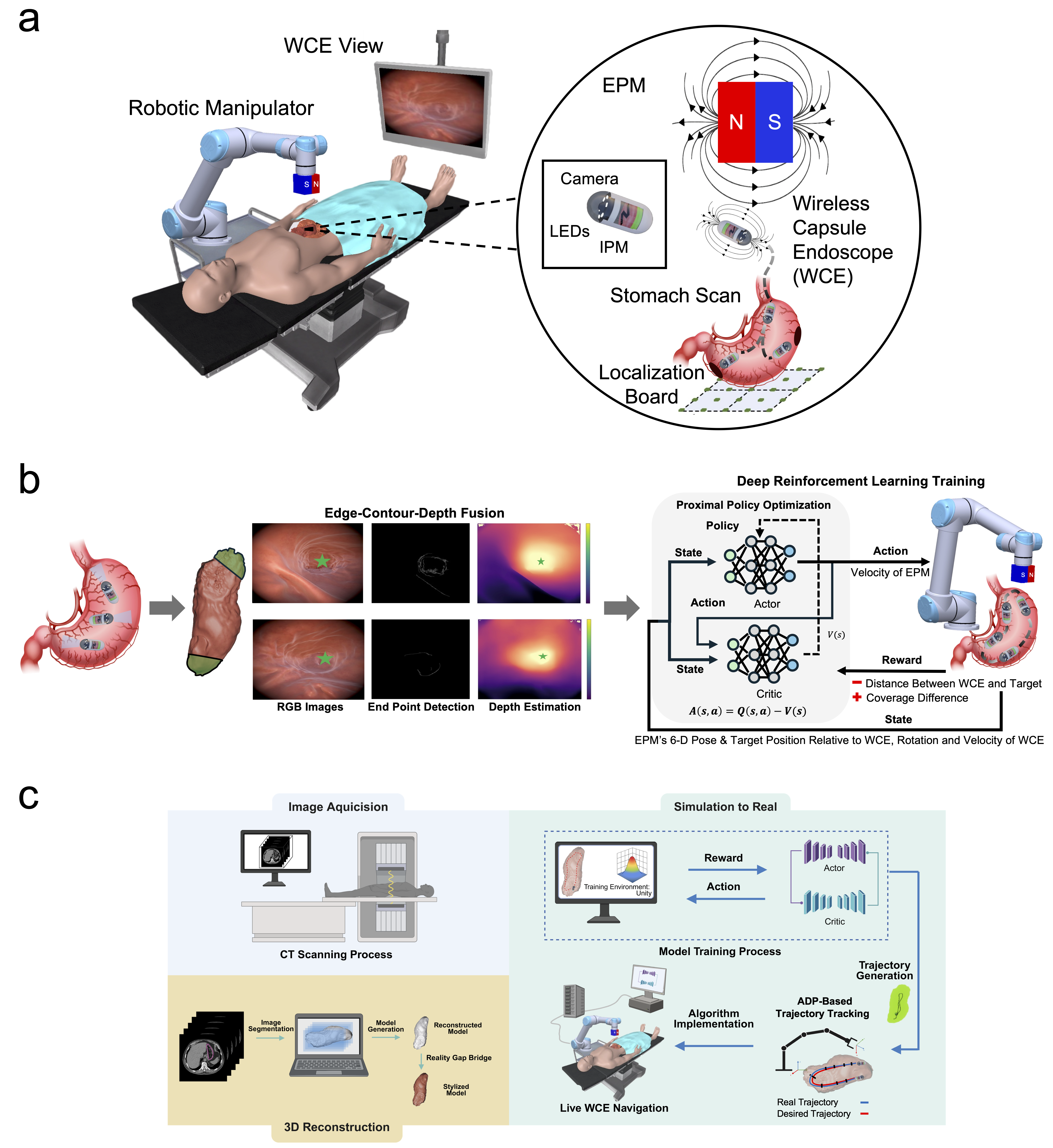}
    \caption{Experimental system and workflow for transferable DRL-based endoscopic navigation. (a)~Schematic of the experimental setup with a single camera capsule inside the stomach and an external 6-DOF robotic arm actuated via a permanent magnet. (b)~AL-DRL training pipeline comprising image acquisition, preprocessing, and policy learning via PPO. (c)~Simulation-to-reality transfer pipeline illustrating the training of a DRL agent in a simulated environment and subsequent deployment in a real-world WCE procedure.}
    \label{fig:fig1}
\end{figure}

To address these challenges, we present a unified framework integrating principled anatomical landmark selection, DRL, and a robust two-stage sim-to-real pipeline. As illustrated in the workflow (Fig.~\ref{fig:fig1}), our approach begins with image acquisition and 3D reconstruction, where patient-specific CT scans are segmented to generate a high-fidelity digital twin. Based on the insight that gastric anatomical structures are highly conserved across patients despite variable mucosal imagery~\cite{dinatale2023}, we replace volatile visual data with stable, low-dimensional landmark coordinates via a lightweight edge-contour-depth fusion module~\cite{ozyoruk2021}. To bridge the dynamic gap, an adaptive dynamic programming (ADP) controller provides robust trajectory tracking by actively mitigating physical disturbances, including actuator latency and peristalsis~\cite{su2023}.

\begin{figure*}[!t]
    \centering
    \includegraphics[width=0.95\textwidth]{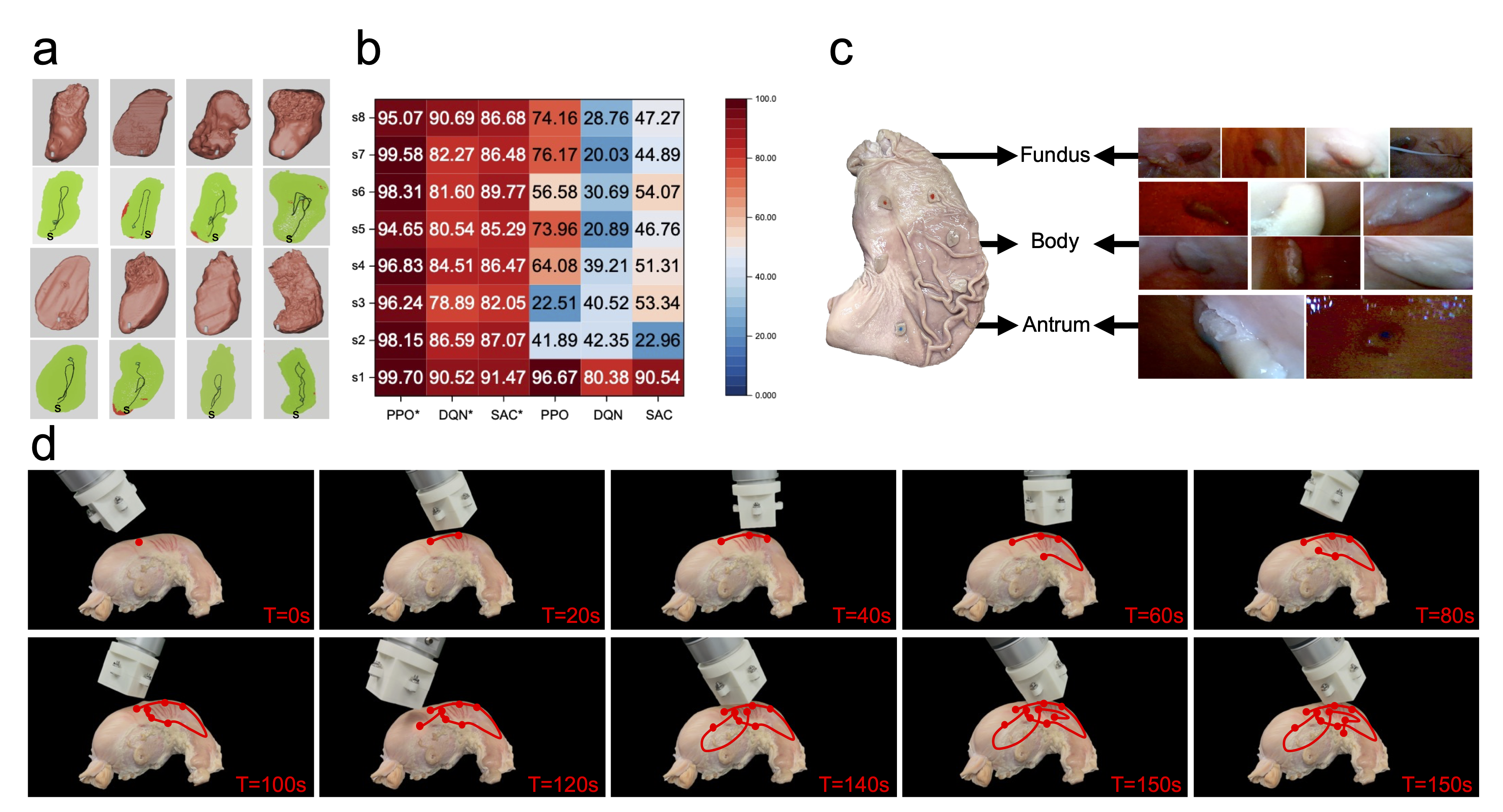}
    \caption{Quantitative results and ex-vivo validation. (a)~Coverage ratio and elapsed time across eight stomach models for vanilla PPO vs.\ AL-DRL. (b)~Mean squared tracking error comparison among ADP, fuzzy, and PID controllers. (c)~Autonomous scanning trajectory on an ex-vivo pig stomach with timestamps.}
    \label{fig:results}
\end{figure*}

By enabling such a high degree of autonomy, this framework effectively overcomes the ``Expert Gap'' prevalent in underserved regions where a shortage of trained endoscopists limits screening capacity. Our system achieves $>97\%$ gastric coverage autonomously by requiring only a standard CT scan for initialization. This streamlined workflow eliminates the need for intensive intraoperative expert intervention, effectively decoupling diagnostic quality from the availability of on-site specialists and establishing a promising paradigm for autonomous gastric navigation.

\section{Methods}

\subsection{Anatomical Landmark-Guided DRL}

The AL-DRL framework adopts a monocular camera-equipped WCE actuated by a 6-DOF robotic arm (UR5) with an external permanent magnet (EPM). The perception module integrates an offline edge-contour-depth fusion pipeline combining Canny edge detection, Hu moment-based contour matching, and lightweight monocular depth estimation (DispNet, 14.84\,M parameters, RMSE 0.37\,cm), achieving 100\% landmark detection across all eight patient-derived anatomies with only 10\,ms per image pair.

Landmark selection is formalized through three criteria: \textit{universality} (conserved across anatomies), \textit{distinction} (perceptually salient), and \textit{navigation utility} (anchoring efficient trajectories). The fundus and pyloric antrum were identified as the optimal pair with 100\% detection accuracy across all models.

The DRL policy (PPO) operates on low-dimensional state vectors comprising 3D landmark coordinates and capsule/EPM 6-DOF poses, bypassing high-dimensional imagery entirely. The reward function is:
\begin{equation}
r_t = k_1 (C_t - C_{t-1}) - k_2 D_t + R
\end{equation}
where $C_t$ is coverage ratio, $D_t$ is distance to the target landmark, and $R$ rewards endpoint traversal.

\subsection{Sim-to-Real Transfer via Hybrid Control}

To bridge sim-to-real discrepancies (actuator latency $\sim$30\,ms, singularities, localization drift), we employ: (1)~a patient-specific digital twin from CT data for offline trajectory planning, and (2)~an adaptive dynamic programming (ADP) controller for online tracking with real-time correction, achieving MSE of 0.04\,cm$^2$---an 89.5\% reduction vs.\ PID control.

\section{Results}

\textbf{AL-DRL performance.}
With landmark guidance, PPO achieved 97.3\% mean coverage (SD 2.0\%) across all eight anatomies---53.8\% coverage increase and 45.3\% time reduction over vanilla PPO (Fig.~\ref{fig:results}a). SAC and DQN improved by 69.1\% and 123.1\% in coverage, respectively (Fig.~\ref{fig:results}b).

\textbf{Ex-vivo validation.}
Five scans on an ex-vivo pig stomach with 12 biological markers yielded 87\% mean coverage in 139.8\,$\pm$\,10\,s (CV 6.5\%), vs.\ 298.4\,$\pm$\,30\,s for a trained operator---53\% time reduction (Fig.~\ref{fig:results}c, d).

\end{document}